\documentclass[a4, 10 pt, conference]{ieeeconf}  

\IEEEoverridecommandlockouts                              

\usepackage{graphicx} 
\usepackage{balance}

\title{\LARGE \bf
Overview of Dialogue Robot Competition 2023
}

\author{Takashi Minato$^{1,2}$, Ryuichiro Higashinaka$^{3}$, Kurima Sakai$^{1}$, Tomo Funayama$^{1}$,\\ Hiromitsu Nishizaki$^{4}$, and Takayuki Nagai$^{5}$ %
\thanks{$^{1}$Hiroshi Ishiguro Laboratories, ATR, Kyoto, Japan
        {\tt\small \{minato, kurima.sakai, funayama\}@atr.jp}}%
\thanks{$^{2}$Guardian Robot Project, RIKEN, Kyoto, Japan
        {\tt\small takashi.minato@riken.jp}}%
\thanks{$^{3}$Graduate School of Informatics, Nagoya University, Nagoya, Japan
        {\tt\small higashinaka@i.nagoya-u.ac.jp}}%
\thanks{$^{4}$Graduate Faculty of Interdisciplinary Research, University of Yamanashi, Yamanashi, Japan
        {\tt\small hnishi@yamanashi.ac.jp}}%
 \thanks{$^{5}$Graduate School of Engineering Science, Osaka University, Osaka, Japan
        {\tt\small nagai@sys.es.osaka-u.ac.jp}}%
}

\begin{document}

\maketitle
\thispagestyle{empty}
\pagestyle{empty}

\begin{abstract}
We have held dialogue robot competitions in 2020 and 2022 to compare the performances of interactive robots using an android that closely resembles a human. In 2023, the third competition DRC2023 was held. The task of DRC2023 was designed to be more challenging than the previous travel agent dialogue tasks. Since anyone can now develop a dialogue system using LLMs, the participating teams are required to develop a system that effectively uses information about the situation on the spot (real-time information), which is not handled by ChatGPT and other systems. DRC2023 has two rounds, a preliminary round and the final round as well as the previous competitions. The preliminary round has held on Oct.27 -- Nov.20, 2023 at real travel agency stores. The final round will be held on December 23, 2023. This paper provides an overview of the task settings and evaluation method of DRC2023 and the results of the preliminary round. 
\end{abstract}

\section{INTRODUCTION}
Voice interactive devices used in our daily lives have been developed from the automated answering services for telephones in the past to the voice agents and AI speakers of smartphones in recent years, and humanoid interactive robots have become the focus of developing next-generation voice interactive devices \cite{Inoue2016}.
For service tasks that require hospitality, a humanoid robot capable of multimodal recognition and expression is particularly effective \cite{Collins2020}. As for the dialogue capability, it has become relatively easy to construct a dialogue system that can sustain a dialogue without breaking down, since dialogue systems using large-scale language (LLM) models, such as GPT \cite{Ham2020}, are now available for everyone to use. However, research is still in progress on how to effectively use the multimodal input/output information of robots for dialogue. Furthermore, for dialogue in the real world, how to effectively use information about the situation on the spot (real-time information), which is not handled by LLM based dialogue systems, is also an issue to be studied.

We have therefore conducted a dialogue robot competition in which robots compete on their dialogue performance in a certain task to encourage researchers to work on dialogue robot tasks and to provide an opportunity to advance the technology by competing in the performance \cite{DRC2020_2022}. This is the world's first attempt to compete in dialogue performance of android robot. So far, two competitions have been held for a dialogue task at a travel agency. This paper outlines the content and the result of preliminary round of the 3rd dialogue robot competition (DRC2023) held in 2023.

DRC2023 followed up on previous competitions with the dialogue task in a travel agency. We performed a two-stage evaluation as done in the previous competition: first, a preliminary round in which the systems of the participating teams were evaluated by the general public, and next, a final round in which the systems were evaluated by designated dialogue researchers to evaluate the technical aspects and by experts working in the tourism industry to evaluate the customer service performance. 

DRC2023 is more advanced than previous competitions in four points.
\begin{itemize}
\item The use of a large-scale language model was recommended.
\item The dialogue task was made more difficult.
\item The evaluation was conducted in a real travel agency store to make the evaluation more practical.
\item The participating teams were provided with the dialogue corpora and recognition systems, that were developed in our research project of "Communicative intelligent systems towards a human-machine symbiotic society."
\end{itemize}
In DRC2022, the only one team used a LLM and that team received the best performance award. In other words, a LLM significantly made the difference in performance. At that time, it was difficult for anyone to use a LLM, but with the appearance of ChatGPT, it became possible for everyone to use LLMs. Everyone is now developing systems based on the assumption of dialogue performance with LLMs. In addition, the use of LLM made it easier to tackle more advanced tasks, so we set up more complex tasks than previous competitions. In the previous competitions, the developed dialogue system were evaluated by setting up a simulated travel agency in the preliminary round. We decided to conduct the evaluation in actual travel agency stores with the cooperation of JTB Corporation and JTB Publishing Inc. for more practical evaluation. We also provided the participating teams with dialogue corpora and recognition systems developed in our research project to support their development while promoting the application of the project's achievements. The preliminary round was held at two travel agencies on Oct.27--Nov.20, 2023. This paper gives an outline of DRC2023 and the results of the preliminary round.

\section{TASK SETTINGS}
The task in the previous competition was to help customers decide on a single sightseeing spot, but in DRC2023, the task is designed to help customers plan to visit multiple sightseeing spots. This task, named the travel itinerary planning task, requires the ability to listen to the customer's requests, to make proposals that meet those requests, and to make feasible plans. Specifically, the customer's purpose is to visit Kyoto City for fun, so the customer decides on a travel plan to visit two sightseeing spots in the city in one day through dialogue. 

The participating teams were provided with information about the sightseeing spots (JTB's "Rurubu DATA"), but they were also allowed to use external resources such as the Web for more informative utterances. The duration of each dialogue was set to approx. 10 minutes, and it was conducted in Japanese. 

\begin{figure}[tb]
    \centering
    \includegraphics[scale=0.5]{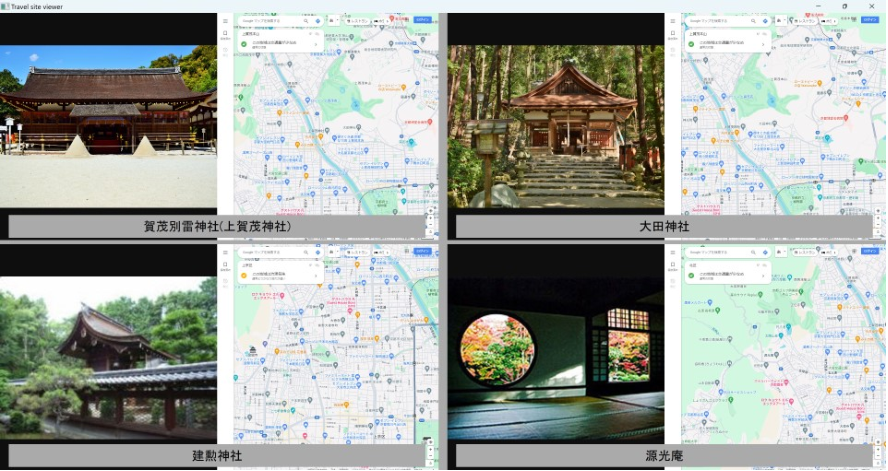}
    \caption{An example of display of sightseeing spots on the monitor}
    \label{fig:monitor}
\end{figure}

The participating teams could use a monitor placed next to the robot to display pictures of the sightseeing spots and maps showing the locations of the sightseeing spots (Fig.~\ref{fig:monitor}). The maximum number of sightseeing spots that can be displayed on the screen at one time is four. The photos of sightseeing spots can be only those included in the "Rurubu DATA". It also contains the latitude and longitude of the sightseeing spots, and the participating teams can access Google Maps to display maps around the sightseeing spots using this information.

\section{AVAILABLE RESOURCES}
\begin{figure}[tb]
    \centering
    \includegraphics[scale=0.105]{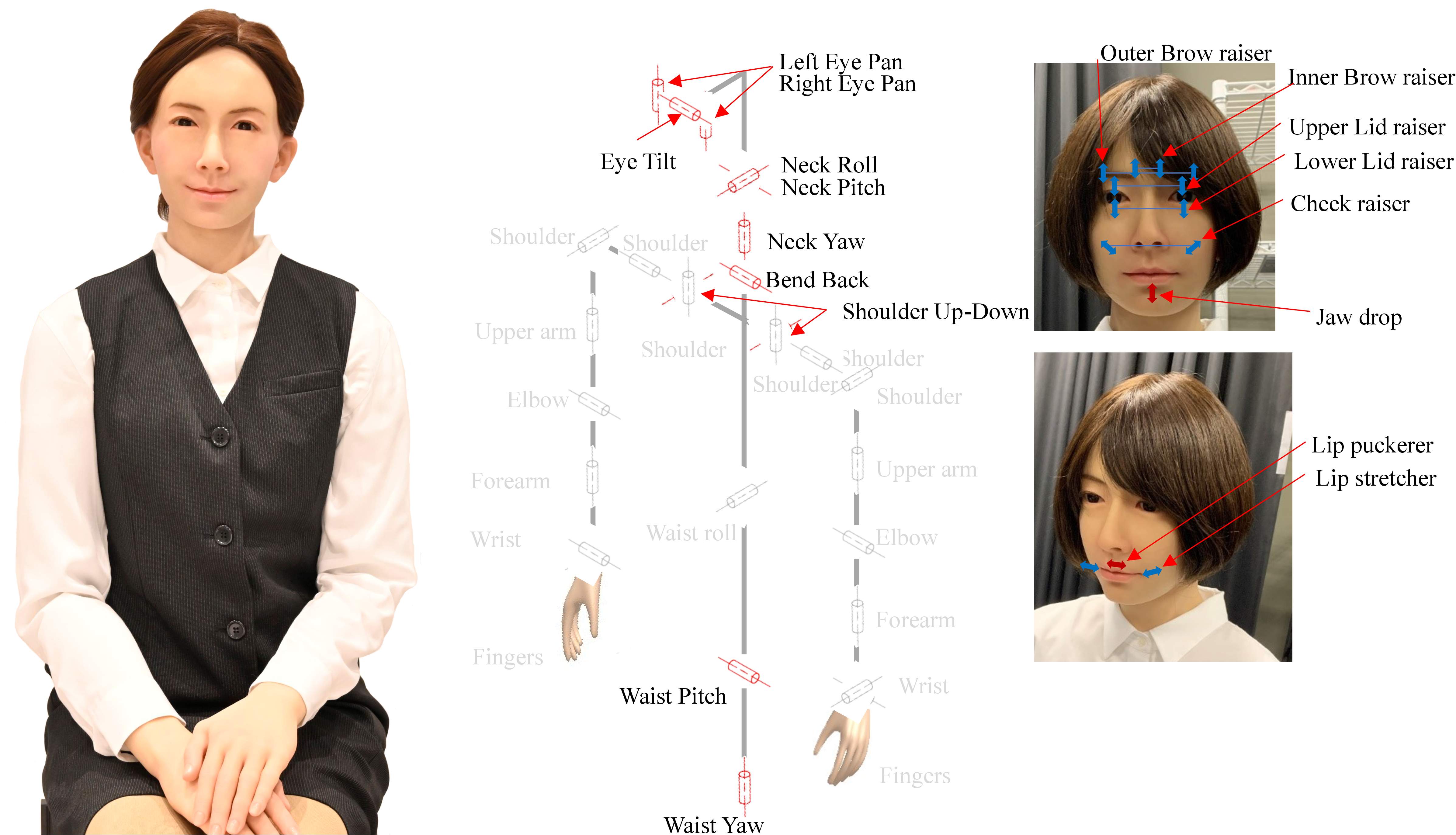}
    \caption{Android I used in the competition}
    \label{fig:android}
\end{figure}

The Android I and Android U were made available as hardware. They have almost same specification. They are approximately 165 cm tall and have a feminine appearance. The face and hands are covered with soft silicone skin. Their degrees of freedom are 18 and 17, respectively (Fig.~\ref{fig:android} shows the controllable joints of Android I) and are capable of lip movements synchronized with speech, head movements such as nodding, eye blinking, gaze behavior, facial expressions, and postural changes of the upper body, all driven by pneumatic actuators. Neither of the arms is movable. 

\begin{figure}[tb]
    \centering
    \includegraphics[scale=0.09]{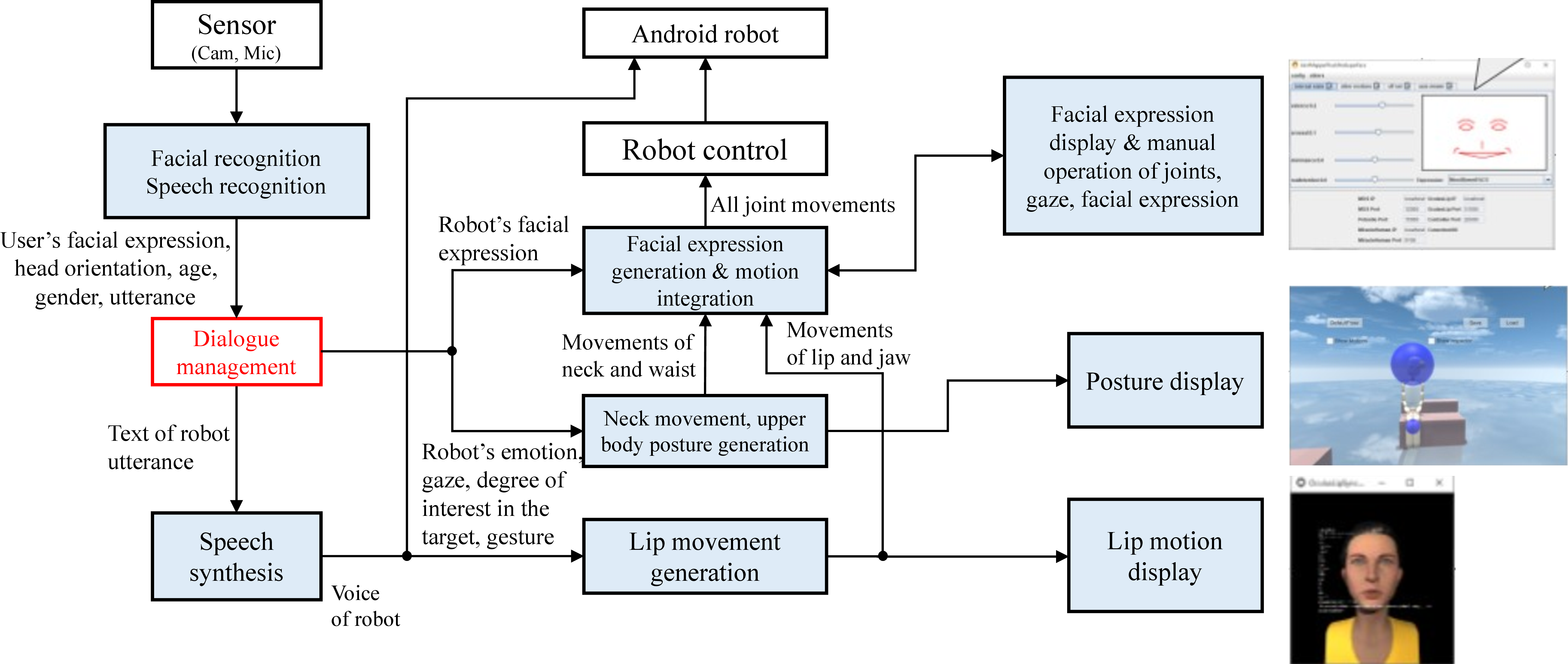}
    \caption{Overview of the middleware distributed to participating teams}
    \label{fig:middleware}
\end{figure}

The participating teams were provided with middleware that integrates face recognition, speech recognition and synthesis, robot facial expression generation, gaze control, lip movement generation \cite{LipSync}, head movement generation \cite{HeadSync}, and posture control (Fig.~\ref{fig:middleware}). The dialogue management module developed by the participating teams simply inputs the customer's facial recognition results (head orientation, facial expression, gender and age), along with the content of the customer's speech, and outputs utterances with the robot's multimodal expressions such as facial expressions, gestures, and body movements as additional information. The middleware comes with a robot simulator, allowing development without the need for an actual robot. It is allowed that the participating teams develop their own voice and image recognition system. A remote testing environment has also been developed so that the participating teams can remotely verify the actual behavior of the robot.

In addition, the participating teams were provided with several module softwares developed in our research project, which support their development of dialogue system. 
\begin{itemize}
\item An emotion estimation program \cite{Katada2022,Komatani2021}\\
This provides the degree to which the customer enjoys interacting with the android based a video data on one turn of interaction between the customer and the android.

\item A program to estimate a person's age and gender from walking video \cite{Shehata2023}\\
This estimates a customer's age and gender based on his gait data observed by Azure Kinect.

\item A program for estimating dialogue acts \cite{inaba-etal-2022-collection}\\
This estimates the dialogue act (intention of speech, such as whether the customer is asking a question, confirming, or agreeing) from the content of the customer's speech.

\item Travel engities generator program\\
This provides an entity to search the "Rurubu DATA" for information needed by a customer based on the content of a dialog between the android and the customer.
\end{itemize}

\section{PRELIMINARY ROUND}
\begin{figure}[tb]
    \centering
    \includegraphics[scale=0.09]{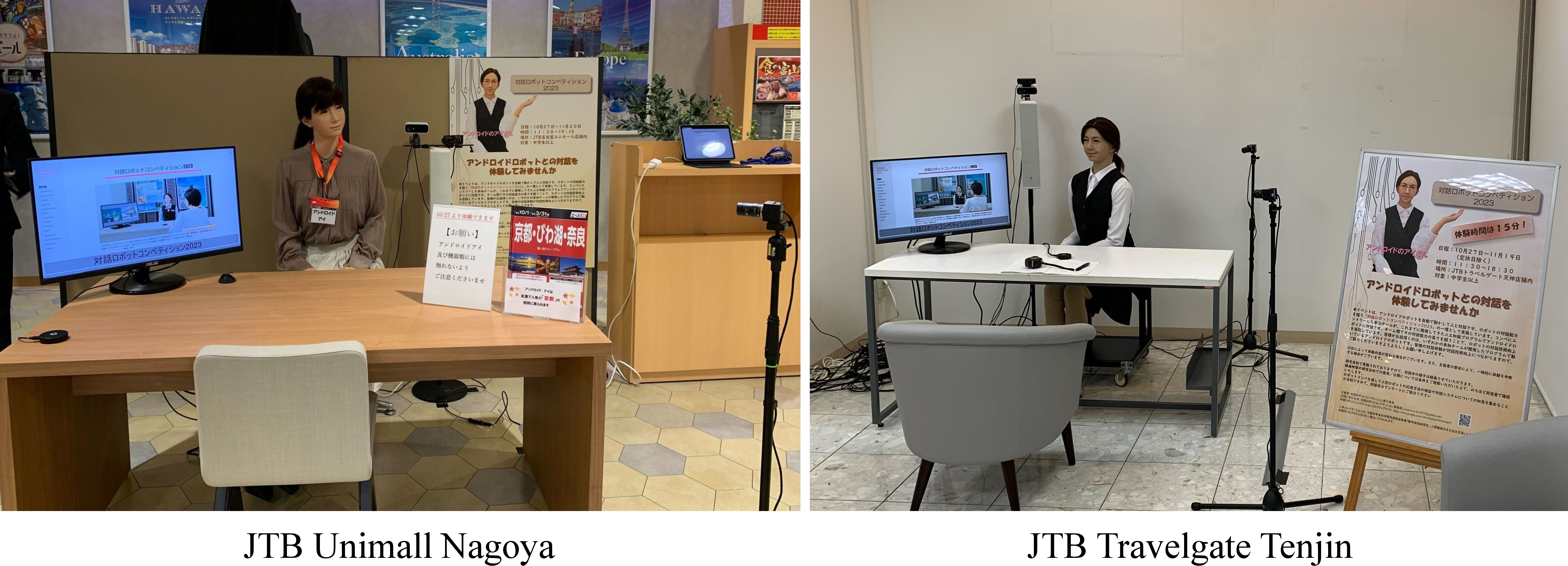}
    \caption{Preliminary round at JTB stores}
    \label{fig:settings}
\end{figure}

The preliminary round was held Oct.27--Nov.20, 2023. For a more practical evaluation, a counter desk for the androids was set up in an actual travel agency (JTB store) (Fig.~\ref{fig:settings}), and the customers interested in talking with the robot had conversation with it while having an instructed mind (considering the content of the dialogue, customers were limited to 12 years of age or older). To obtain evaluation from many customers, we installed Android U and Android I at two locations (JTB Unimall Nagoya (Nagoya city) and JTB Travelgate Tenjin (Fukuoka city)). The participating teams evaluated the developed system for one day at each store. 

The customers who wished to talk with the robot received an explanation of the dialogue situation from a staff member. The situation is that the purpose of the customer is to visit a certain city (Kyoto city) for fun, and to decide on a travel plan to visit two sightseeing spots in the city in one day, while interacting with the robot. The maximum duration of the dialog is 10 minutes, and the robot must end the dialog after 10 minutes.

The dialogue performance was evaluated by two factors: impression evaluation and feasibility of the decided travel plan. The impression evaluation consisted of the following nine questions (originally in Japanese; translated here into English). Each item was measured on a 7-point Likert scale. 
\begin{itemize}
    \item Informativeness: "Were you able to obtain sufficient information about the sightseeing spots?"
    \item Naturalness: "Did you have a natural dialogue with the robot?"
    \item Appropriateness: "Was the robot's service appropriate?"
    \item Likeability: "Was the robot likable in providing the service?"
    \item Satisfaction with dialogue: "Were you satisfied with your interaction with the robot?"
    \item Trustworthiness of robot: "Did you trust the robot?"
    \item Usefulness: "Did you use the information obtained from the robot to select the sightseeing spot?"
    \item Trustworthiness of provided information: "Did you trust the information provided by the robot?"
    \item Intention to reuse: "Would you like to visit this travel agency again?"
\end{itemize}

To evaluate the feasibility of the plan, we asked the following two questions (originally in Japanese; translated here into English) to the customers. Each item was answered with a yes or no response.
\begin{itemize}
    \item Plan making: Have you been able to make plans to visit two sightseeing spots?
    \item Plan feasibility: Do you think the plan is feasible (judging by your knowledge)?
\end{itemize}

To comprehensively evaluate the two factors of impression and feasibility, the two evaluation scores (total of averaged impression scores and rate of customer who answered yes for both plan making and plan feasibility questions) for each participating team were plotted on a scatter plot with two axes. In this plot, a team belonging to the cluster formed at the position with the highest values on both axes was considered a top team of preliminary round and advanced to the final round. The teams belonging to the cluster formed at the second highest position received an honorable mention award.

\begin{figure}[tb]
    \centering
    \includegraphics[scale=0.17]{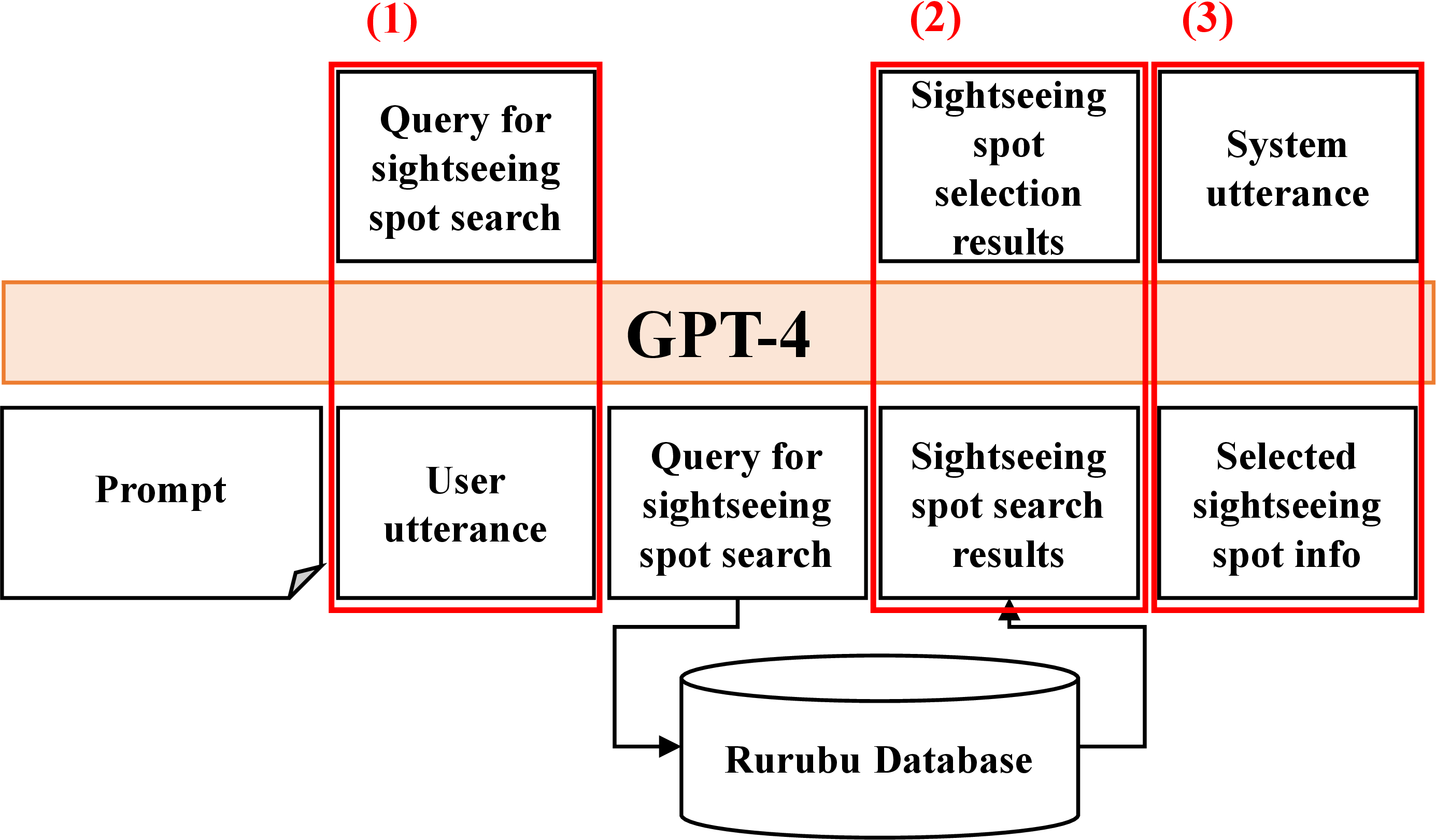}
    \caption{Overview of baseline system}
    \label{fig:baseline}
\end{figure}

The baseline system uses GPT-4 \cite{GPT-4}, a LLM developed by OpenAI, to generate sightseeing spot search queries, select sightseeing spots to be included in the utterances, and generate system utterances. Figure~\ref{fig:baseline} shows the inputs and outputs of the language model in the baseline system. When the system receives a user utterance, it first generates a sightseeing spot search query using the language model (Fig.~\ref{fig:baseline} (1)). It outputs the name of the sightseeing spot, the attributes of the sightseeing spot, the area where the sightseeing spot is located, the time required for sightseeing, and the recommended season for sightseeing as the result of the generated query, such as \{"area": ["Higashiyama-ku"], "category": ["sightseeing-shrine/temple/church", "sightseeing-building/historical site"]\}. Next, the generated query is input to the "Rurubu DATA" Web Service to obtain information on sightseeing spots that match the conditions. The language model selects sightseeing spots to be included in the utterance from the list of sightseeing spot names included in the search results  (Fig.~\ref{fig:baseline} (2)). It selects no more than three sightseeing spots. Finally, the language model generates an utterance from detailed information on the selected sightseeing spots  (Fig.~\ref{fig:baseline} (3)). In each of the processes, the input to the language model consists of a prompt, a history of sightseeing spot search query generation, sightseeing spot selection, and utterance generation. The prompts describe the flow of the dialogue and instructions for generating sightseeing spot search queries, selecting sightseeing spots, and generating utterances.

\begin{figure}[tb]
    \centering
    \includegraphics[scale=0.3]{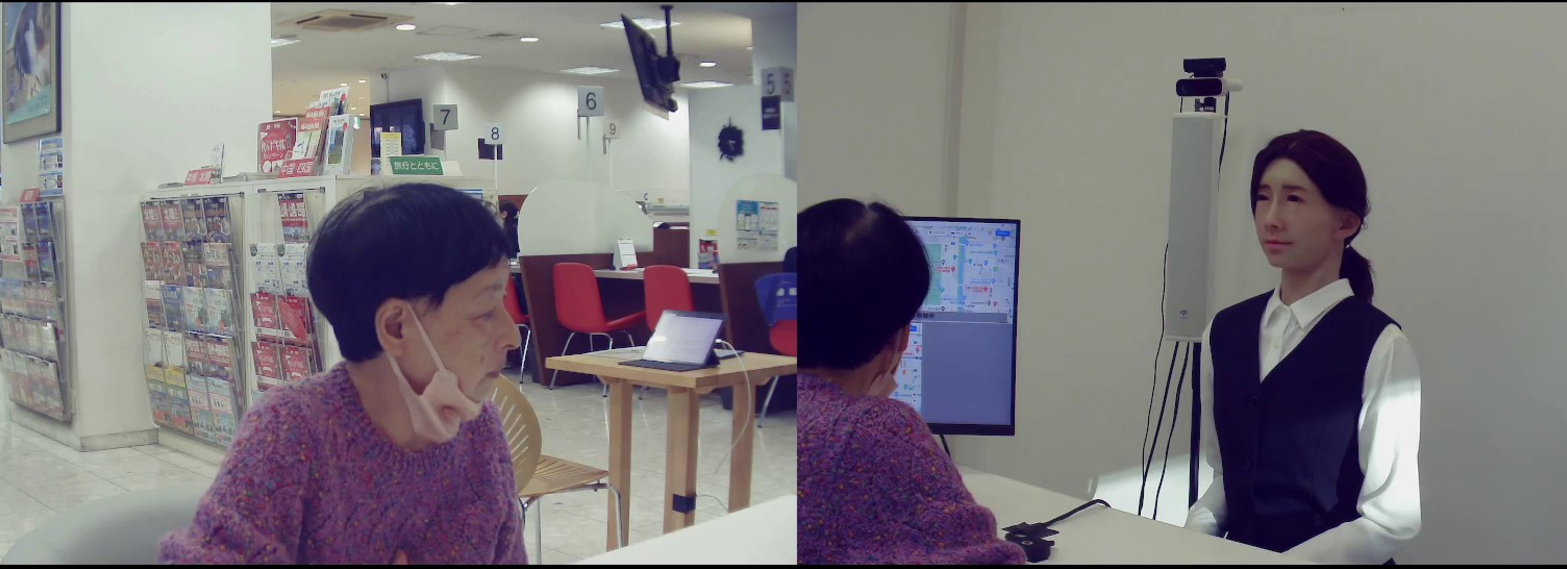}
    \caption{Example scene of dialogue in the preliminary round (Tenjin store)}
    \label{fig:scene}
\end{figure}

\begin{figure}[tb]
    \centering
    \includegraphics[scale=0.20]{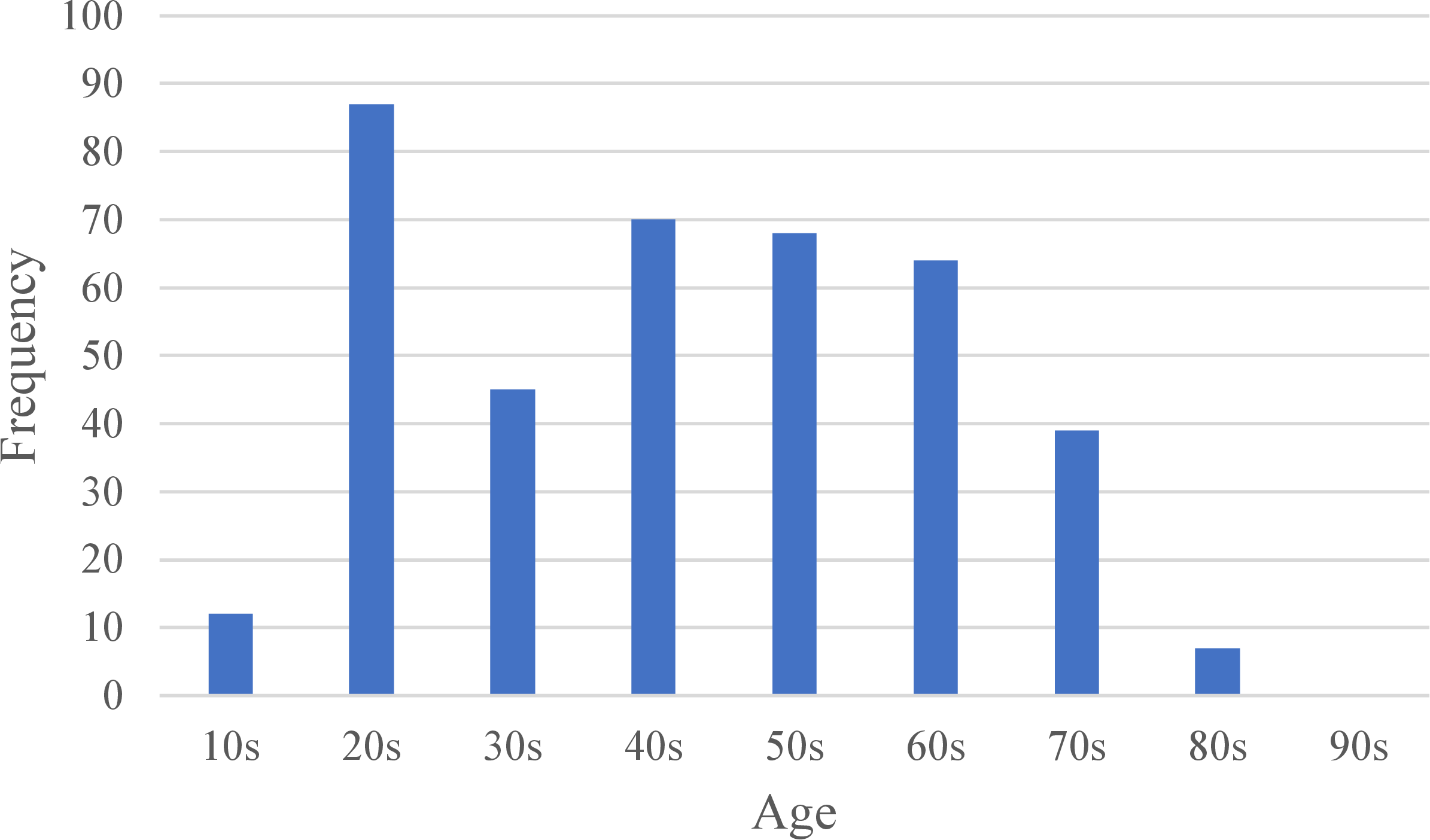}
    \caption{Age distribution of customers}
    \label{fig:age}
\end{figure}

Nineteen teams submitted entries to DRC2023, and finally twelve teams (DSML-TDU, Flow, irisapu, JERRY\_cis, MIYAMA, Nit,Oita, NTT-EASE, ROS, SBIntuitions, Tokuolab, TOM\_cis, UEC-IL) participated in the preliminary round along with the baseline system. Eight of the teams were from universities, two from a college of technology, and two from industry. A total of 419 customers evaluated the system in the preliminary round. Figure~\ref{fig:scene} shows a scene of a customer talking with the robot. The average number of customers handled per team was 18.5 except for baseline (the number of customer for baseline was more than the other teams since baseline system was evaluated in multiple days in both stores). From the age distribution of customers shown in Fig.~\ref{fig:age} (some customers did not answer their age), it can be seen that a wide range of generations, from teens to people in their 80s, talked with the robot. 

\begin{figure}[tb]
    \centering
    \includegraphics[scale=0.1]{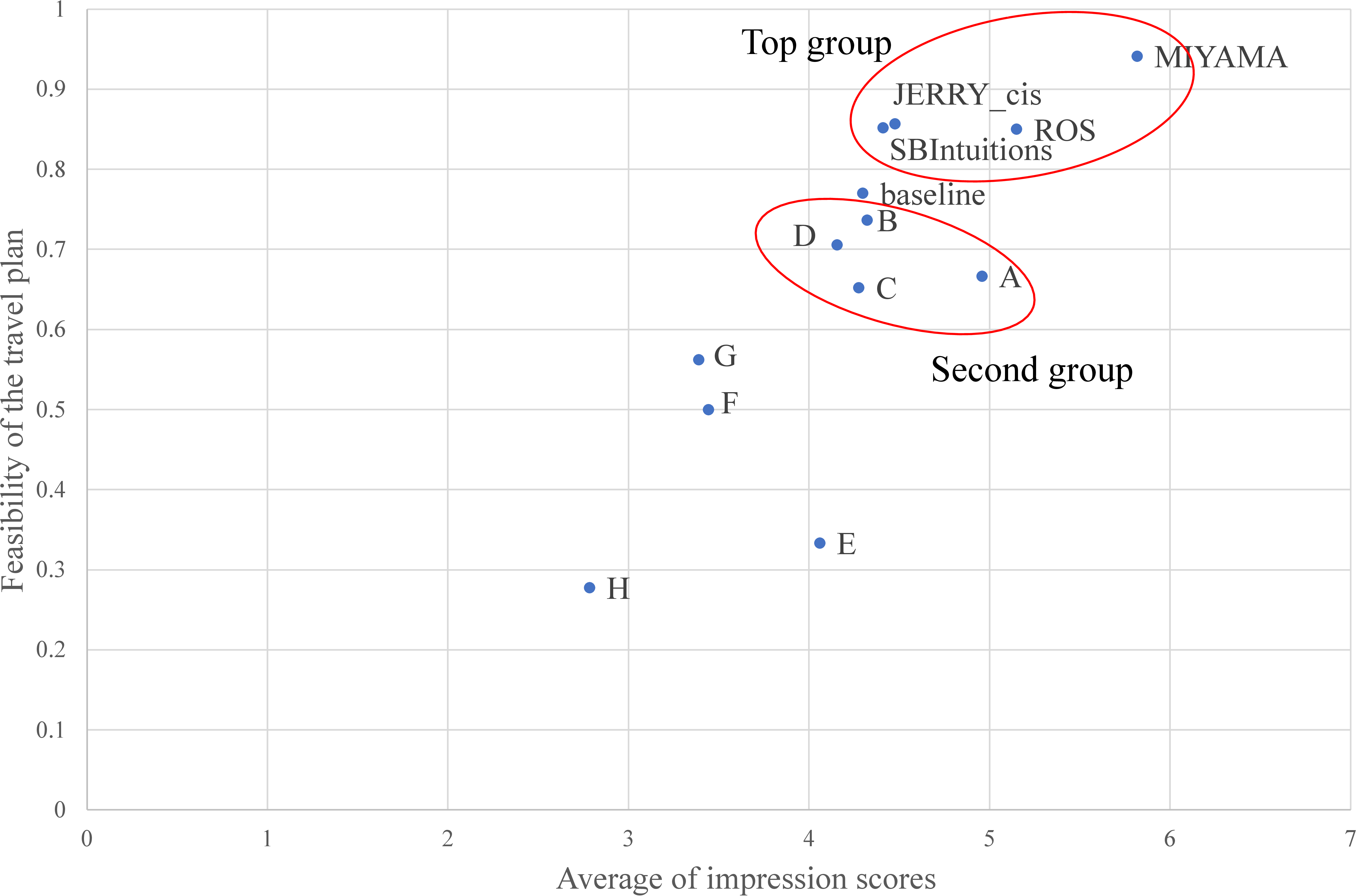}
    \caption{Scatter plot with impression and feasibility scores}
    \label{fig:plot}
\end{figure}

The evaluation results are shown in Table~\ref{table:result}, and the scatter plot is shown in Fig.~\ref{fig:plot}. Teams that did not make it to the final round are anonymized. Since the scores of four teams (MIYAMA, ROS, JERRY\_cis, and SBIntuitions) significantly outperformed those of the other teams and baseline in both factors, these four teams were selected as finalists.  

\begin{table*}[tb]
    \caption{Results of preliminary round. The teams are sorted by total impression score. Inf, Nat, App, Lik, Sat, Tru/r, Use, Tru/i, Reu, Ave imp, Fea denote Informativeness, Naturalness, Appropriateness, Likeability, Satisfaction with dialogue, Trustworthiness of robot, Usefulness, Trustworthiness of provided information, Intention to reuse, Average of impression score, and Feasibility of the plan, respectively.}
    \label{table:result}
    \begin{tabular}{c} 
    \begin{minipage}{180mm}
      \centering
      \scalebox{0.24}{\includegraphics{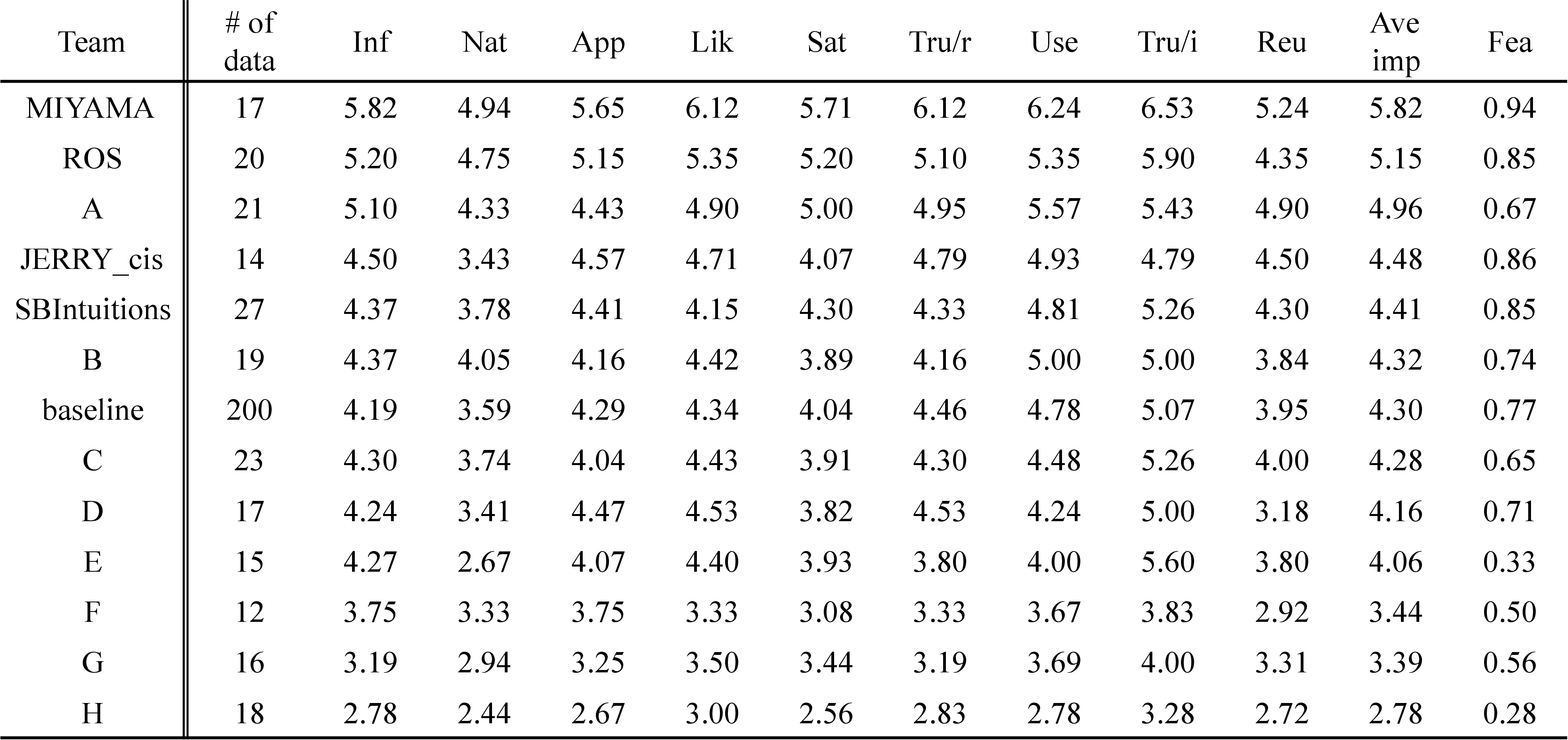}}
    \end{minipage} 
    \end{tabular}
\end{table*}

There is a strong correlation between the scores of the two factors. However, some teams had a low score for only one of the two factors, suggesting that the two factors contributed to the evaluation of the overall performance. Teams A, B, C, and D form the second group in Fig.~\ref{fig:plot}. Those four teams received an honorable mention award. 

\section{OVERVIEW OF DIALOGUE SYSTEMS DEVELOPED BY PARTICIPATING TEAMS}

In this competition, all teams developed systems using LLM. Therefore, the usage of LLMs may have made the difference in the evaluation among the teams. Several teams used multiple LLMs to understand the context of the dialogue and to understand the intentions of the customers. In addition, several teams used routing services to improve the feasibility of travel plans. In utterance generation using LLMs, the response time of the robot becomes longer in real-time dialogue tasks. The efforts to shorten the response time and to make the customers feel the response time is not long had a significant effect on the impression score. For the dialogue robot competition, it is desirable to come up with a variety of ideas for improving non-verbal interaction. In the impression evaluation, team MIYAMA received a higher score than the other teams. This team included dialogue praising the customer based on the customer's appearance, and it is thought that the design of speech that makes rapport with the customers based on real-time information contributed greatly to the improvement of the impression. Face-to-face dialogue service, which requires hospitality, requires both dialogue tasks and relationship building with the customer, and creating dialogue tasks that satisfy both will lead to the development of practical dialogue robot technology. The following is a brief description of the characteristics of each team.

\noindent
{\bf Team DSML-TDU} \cite{DRC2023_DSML-TDU} They used two types of LLM prompts: prompts to control the flow of the dialogue and prompts to estimate cognitive states related to the customer's turn-taking. They used the latter to prevent unintentional interruptions and make smooth and natural dialogue.

\noindent
{\bf Team Flow} \cite{DRC2023_Flow} The preferences related to sightseeing spots are extracted from the customers' utterances by using LLM, and a common ground related to sightseeing spots is constructed using a tree structure model. In addition, for natural turn-taking, they used the provided Travel Agency Task Dialogue Corpus \cite{inaba-etal-2022-collection} to build training data. Based on this, they constructed a model that generates responses and nodding at appropriate timing from the utterances by machine learning method.

\noindent
{\bf Team  irisapu} \cite{DRC2023_irisapu} A dialogue scenario consisting of five phases was designed, and by changing the description of the LLM prompts in each phase, the robot's utterance was generated according to the scenario. In addition, the LLM was used to detect dialogue breakdowns, and recovery utterances were generated when detected.

\noindent
{\bf Team JERRY\_cis} \cite{DRC2023_JERRY_cis} Multiple LLMs were used asynchronously in multiple roles to accelerate dialogue responses. One LLM stores the history of the current prompt utterance, and the other summarizes past conversations. These two are considered to correspond to short-term and long-term memory.

\noindent
{\bf Team  MIYAMA} \cite{DRC2023_MIYAMA} The dialogue strategy consists of a phase of listening to the customer's preferences and experiences and a phase of proposing a travel plan based on these preferences and experiences. The relationship with the customer is established by making a complement based on the apparent information of the customer observed by a camera.

\noindent
{\bf Team Nit,Oita} \cite{DRC2023_NitOita} ChatGPT was used as the LLM-based dialogue system and Function Calling of ChatGPT was used to prevent hallucination.

\noindent
{\bf Team NTT-EASE} \cite{DRC2023_NTT-EASE} The dialogue flow was manually built, and the answers to customers' questions were generated using LLM. The system collected user preferences and recommended a travel plan. At that time, the user's response was evaluated by sentiment analysis. For smooth dialogue, fillers were inserted before the LLM generates a response.

\noindent
{\bf Team ROS} \cite{DRC2023_ROS} The combination of rule-based utterance selection and utterance generation using LLM was used to ensure the quality of robot utterances and to be able to respond to unexpected utterances by the customer. In order to capture the customer's intentions, they constructed a yes/no classification using the provided sentiment estimator \cite{Komatani2021,Katada2022} and BERT so that the customer's intentions can be properly reflected in the plan-making. The system presents accurate information on routes around sightseeing spots using NAVITIME.

\noindent
{\bf Team SBIntuitions} \cite{DRC2023_SBIntuitions} They designed a scenario consisting of multiple interaction phases, and developed a system in that the phase transitions while satisfying the conditions of each phase. They designed a strategy that builds a large number of travel plans in advance and recommends suitable one to reduce the cognitive load of customers in their decision-making process. They used NAVITIME to provide accurate information on routes around sightseeing spots.

\noindent
{\bf Team Tokuolab} \cite{DRC2023_Tokuolab} The system uses an LLM-based dialogue system to respond to the customers. The robot recognizes the customer's facial expressions through camera image, and sympathetically behaves by expressing the same facial expressions.

\noindent
{\bf Team  TOM\_cis} \cite{DRC2023_TOM_cis} Multiple LLMs were used asynchronously in multiple roles to accelerate the response of the system in the dialogue. One LLM was used to provide an appropriate response and the other LLM was used to understand the customer's intentions and search the information which matches the customer's intention in the database. NAVITIME was used to provide accurate information on routes around sightseeing spots.

\noindent
{\bf Team  UEC-IL} \cite{DRC2023_UEC-IL} By creating a dialogue scenario and inputting only the necessary dialogue context and sightseeing spot information into the model according to the dialogue scenario, the flow of dialogue is controlled and consistent responses are generated. With respect to nonverbal expressions, the customer's satisfaction was improved by introducing behavior and speech control that emphasizes important words according to the robot's utterance and the customer's state.

\section{FINAL ROUND}
The final round will be held on December 23, 2023 at the National Museum of Emerging Science and Innovation (Miraikan). In the final round, the systems will be evaluated by designated dialogue researchers and by experts working in the tourist industry. The dialogue will be open to researchers watching.

\section{CONCLUSION}
This paper provided an overview of DRC2022 and the results of the preliminary round. The top-four teams selected as finalists significantly outperformed the other teams in two factors: impression and plan feasibility. The two factors could evaluate the overall performance of both the impression of the dialogue and the dialogue task.

\balance




\section*{ACKNOWLEDGMENT}

This work was partially supported by a Grant-in-Aid for Scientific Research Grant Number JP19H05692 (managing the competition, data analysis, and writing paper) and JST Moonshot R\&D Grant Number JPMJMS2011 (conducting the preliminary round at the travel agencies).
We thank Prof. Ogawa of Nagoya University for providing the Android U for the preliminary round.


\bibliographystyle{IEEEtran}
\bibliography{drc2023}

\end{document}